\documentclass{article}
\usepackage{spconf,amsmath,epsfig}
\usepackage{amsfonts}
\usepackage{hyperref}
\usepackage{cleveref}
\usepackage{graphicx}
\usepackage{tabularx}
\usepackage{pythonhighlight}
\let\OLDthebibliography\thebibliography
\renewcommand\thebibliography[1]{
  \OLDthebibliography{#1}
  \setlength{\parskip}{0pt}
  \setlength{\itemsep}{0pt plus 0.3ex}
}

\begin{document}\sloppy

\title{IEEE ICME 2024 Grand Challenge: Low-power Efficient and Accurate Facial-Landmark Detection for Embedded Systems}
%
\name{Zong-Wei Hong, Yu-Chen Lin}
\address{}

\maketitle

\begin{abstract}
   The domain of computer vision has experienced significant advancements in facial-landmark detection, becoming increasingly essential across various applications such as augmented reality, facial recognition, and emotion analysis. Unlike object detection or semantic segmentation, which focus on identifying objects and outlining boundaries, facial-landmark detection aims to precisely locate and track critical facial features.
   
   However, deploying deep learning-based facial-landmark detection models on embedded systems with limited computational resources poses challenges due to the complexity of facial features, especially in dynamic settings. Additionally, ensuring robustness across diverse ethnicities and expressions presents further obstacles. Existing datasets often lack comprehensive representation of facial nuances, particularly within populations like those in Taiwan.

    This paper introduces a novel approach to address these challenges through the development of a knowledge distillation method. By transferring knowledge from larger models to smaller ones, we aim to create lightweight yet powerful deep learning models tailored specifically for facial-landmark detection tasks. Our goal is to design models capable of accurately locating facial landmarks under varying conditions, including diverse expressions, orientations, and lighting environments. The ultimate objective is to achieve high accuracy and real-time performance suitable for deployment on embedded systems. This method was successfully implemented and achieved a top 6th place finish out of 165 participants in the IEEE ICME 2024 PAIR competition.
\end{abstract}
\begin{keywords}
  Facial landmark detection, Deep learning, Low power
\end{keywords}
\section{Introduction}

To improve the accuracy of facial landmark detection using a \textit{single model}, we employ a two-step approach, dividing the process into face detection and landmark detection. Specifically, we utilize on-the-fly packages such as OpenCV~\cite{opencv_library} or dlib~\cite{6909637} due to time constraints in pre-processing. It is important to note that \textbf{NEITHER} of these packages is based on deep learning techniques.. Another advantage of this two-step approach is that it allows for potential enhancements in accuracy by substituting the face detector with potentially more powerful alternatives in the future.

In our approach, we employ the knowledge distillation technique to enhance the performance of our models. Initially, we trained a Swin Transformer (SwinV2~\cite{liu2022swin}) utilizing the STAR Loss \cite{zhou2023star} function as our teacher model , yielding a promising score of 18.08 in the initial round of experimentation. Subsequently, leveraging the distilled knowledge from SwinV2, we employed it to train a more lightweight model, the MobileViT-v2~\cite{mehta2022separable}. Remarkably, even in its nascent stage, the MobileViT-v2 showcased significant promise by achieving a score of 15.75 in the initial round of evaluation.

We will provide further details in the following section.

\section{Implementation Techniques}
\subsection{Preliminary}
Given a cropped face image $\mathcal{I}$, the main goal of facial landmark detection is to identify and recover the $N$ facial landmarks, denoted as $\mathcal{P} = \{P_i = \{x_i, y_i\}| i = 1, ..., N\}$. This task typically involves two distinct approaches that complement each other: regression-based and heatmap-based methods. 

Due to their superior accuracy, we have opted to incorporate heatmap-based methods into the development of our approach. Specifically, for every landmark, we produce a corresponding heatmap. These normalized heatmaps can be understood as probability distributions representing the predicted facial landmarks. The predicted coordinates are extrapolated from the heatmaps utilizing a soft-Argmax~\cite{nibali2018numerical} decoding technique.

Specifically, given a discrete probability distribution $h$, we assign the value $h_k$  to represent the probability of the predicted landmark being positioned at $o_k \in \mathbb{R}^2$  for each predicted landmark $\hat{P}_i$. This calculation can be performed using the following formula.
\begin{equation}
    \hat{P}_i = \sum_{k}h_ko_k
\end{equation}

Therefore, we can employ the regression loss $\mathcal{L}_{reg}$ to direct the training of our model. To maintain generality, we opt for the STAR loss~\cite{zhou2023star} in our experimental setup.
\begin{figure*}
    \centering
    \resizebox{2\columnwidth}{!}{
    \includegraphics{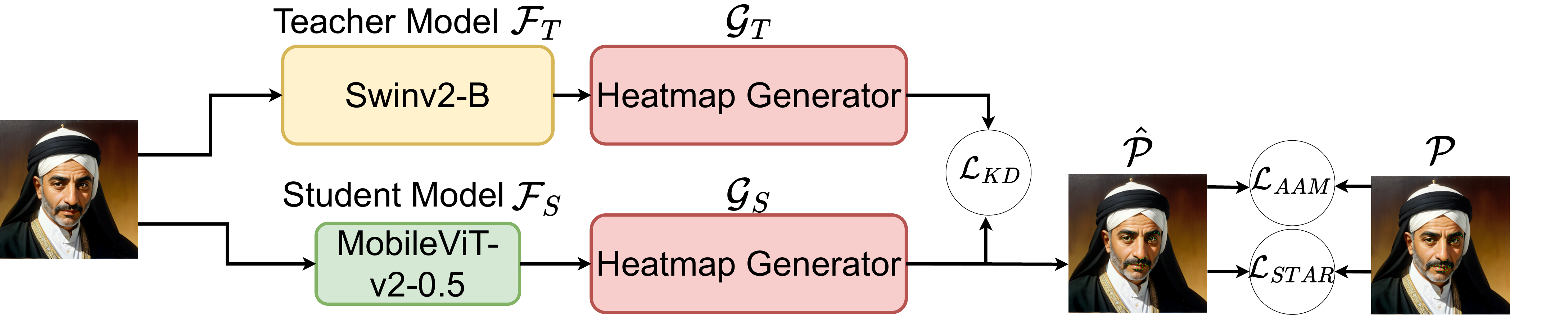}
    }
    \caption{We propose a two-stage training process for our method. In the first stage, we train the teacher model, denoted as $\mathcal{F}_{T}$, using a combination of two loss functions: $\mathcal{L}_{AAM}$ and $\mathcal{L}_{STAR}$. Subsequently, in the second stage, we train the student model by introducing an additional loss function, $\mathcal{L}_{KD}$.}
    \label{fig:enter-label}
\end{figure*}
\subsection{Model Architecture}
There are primilary two components of our model: the feature extractor backbone, denoted as $\mathcal{F}:\mathbb{R}^{3\times256\times256}\rightarrow\mathbb{R}^{C\times64\times64}$, and the heatmap generator, represented by $\mathcal{G}:\mathbb{R}^{C\times64\times64}\rightarrow\mathbb{R}^{N\times64\times64}$, which are illustrated in \Cref{fig:enter-label}. Therefore, we can define the predicting process as follows:
\begin{equation}
    \hat{P}_i = \sum_{k}[\mathcal{G}(\mathcal{F}(\mathcal{I}))]^i_ko_k,
\end{equation}
where $[\mathcal{G(\mathcal{F(\cdot)})}]^i_k$ represents the predicted probability heatmap for the $i$-th facial landmark at location $o_k$. It is noteworthy that, for the purpose of enhancing the heatmap's precision, we have integrated the Anisotropic Attention Module (AAM), as detailed in~\cite{huang2021adnet}. The AAM functions as an attention module, tasked with generating a point-edge heatmap that functions as an attention mask. This module ensures that the heatmap demonstrates qualities akin to a blend of Gaussian distribution and neighboring boundary distribution. During the training phase, we employ $\mathcal{L}_{AAM}$ to guide the edge and point, and we highly encourage readers to refer to the original paper~\cite{huang2021adnet} for comprehensive details.

Recently, it has become evident that transformer-based architectures outperform CNNs. Moreover, given that this challenge expected a low-power, efficient solution for embedded systems, architecture specifically designed for mobile devices such as MobileViT is definitely a good choice. Consequently, this paper adopts the transformer-based architecture as the cornerstone of our features, employing knowledge distillation to refine our approach.
\subsubsection{Teacher architecture}
For the selection of our teacher model, we have chosen SwinV2-B~\cite{liu2022swin}, pre-trained on ImageNet-1K. In our implementation, we have omitted the classifier layer, utilizing only the feature extractor component as the backbone for our teacher model. To ensure compatibility with the input dimensions of the heatmap generator $\mathcal{G}$, we have incorporated an upsample layer to the output of SwinV2.

\subsubsection{Student architecture}
For our student model selection, we've opted for MobileViT-v2-0.5~\cite{mehta2022separable} as our backbone due to its suitability for our embedding requirements. We've specifically chosen to exclude the classifier layer, utilizing solely the feature extractor component as the backbone, and subsequently increasing the feature dimension to 64 through upsampling. However, in comparison to the original version, we have made partial structural revisions to ensure its compatibility with tflite-runtime versions up to 2.11.0.

The specific revision involves replacing the "fold" and "unfold" operations in PyTorch with "reshape" and "permute". However, the "permute" operation does not support 6 dimensions for tflite-runtime versions up to 2.11.0. Therefore, we split the original 6-dimensional input into multiple 5-dimensional tensors, permute each 5-dimensional tensor, and finally concatenate them back into a 6-dimensional tensor. Below is our code (partly pseudo):
\begin{python}
# mobilevit_block.py
def unfolding_pytorch(...):
        ...
    #Original code
    #patches = nn.functional.unfold(
    #    ...
    #)
    feature_map = feature_map.reshape(...)
    x_splits = torch.split(...)
    splited_transposed_tensors = []
    for x_split in x_splits:
        squeezed_tensor = torch.squeeze(x_split, 
        dim=-1
        )
        splited_transposed_tensors.append(
        torch.unsqueeze(
        squeezed_tensor.permute(0, 1, 3, 2, 4), 
        dim=3)
        )
    feature_map = torch.cat(
    splited_transposed_tensors, 
    dim=3
    )
    
def folding_pytorch(...)
    #Original code
    #feature_map = nn.functional.fold(
    #    ...
    #)
    ......
    # Similar strategy here.
    return feature_map

\end{python}
\subsubsection{Heatmap generator architecture}
Our heatmap generator architecture, as outlined by~\cite{huang2021adnet}, employs Convolution2D layers with sigmoid activation for edge and point prediction, alongside Convolution2D layers with instance normalization and ReLU activation for heatmap prediction.
\subsection{Knowledge Distillation Loss}
For training teacher model, we only use $\mathcal{L}_{reg}$ and $\mathcal{L}_{AAM}$ for supervising. 
However, in order to facilitate the learning process for the lightweight model, we have designed the strainforward loss $\mathcal{L}_{KD}$ to efficiently transfer primary features from the teacher model to the student model, defined as:
\begin{equation}
    \mathcal{L}_{KD} = \sum_{i =1}^N\sum_{k}|| [\mathcal{G}_T(\mathcal{F}_T(\cdot))]^i_k - [\mathcal{G}_S(\mathcal{F}_S(\cdot))]^i_k ||_2,
\end{equation}
where $\mathcal{F}_T$ denotes the teacher feature extractor, $\mathcal{F}_S$ denotes the student feature extractor, $\mathcal{G}_T$ denotes the teacher heatmap generator and  $\mathcal{G}_S$ denotes the student heatmap generator.
\section{Experiment}
\subsection{Experiment setting}
All experiments are conducted using the RTX 3090 and Intel i7-12700K configurations.

The data augmentation process employed in~\cite{zhou2023star, huang2021adnet} remains consistent across all experiments. It involves generating the input image through two sequential steps: (1) Cropping the facial regions and resizing them to 256x256 dimensions. (2) Implementing augmentations including random rotation ($45^\circ$), random scaling ($\pm10\%$), random crop ($\pm18\%$), random blur ($40\%$), random gray ($20\%$), random occlusion ($40\%$), and random horizontal flip ($50\%$)

For the training strategy, we have designated a learning rate of 2e-4 for the backbone component and 1e-3 for the heatmap component during both teacher and student training, utilizing the AdamW~\cite{loshchilov2017decoupled} optimizer. The batch size for training both the teacher and student is configured to be 16.

For the dataset, we partitioned the training dataset into a training set and a validation set using a 10:1 ratio and selected the model with the lowest normalized mean error (NME).


\begin{table}[tb!]
\centering
\resizebox{1\columnwidth}{!}{
\begin{tabular}{c|c|c|c|c|c} 
\hline
Method          & GFlops     & MMACs & Params (M) & Inference Time (ms)& Val NME (\%) \\
\hline
Hourglasses  & 10.4436    & 5190.2       & 1.8747           & 88         & 3.78         \\
mobileFormer-26m  & 0.288    & 142.478        & 1.5152      & 31.1         & 5.31         \\
MobileViT-v2-0.5 w/o $\mathcal{L}_{KD}$ & 1.1865   & 581.354  & 1.1419      & 23.8        & 5.13        \\
MobileViT-v2-0.5 (Ours) & 1.1865   & 581.354  & 1.1419      & 23.8        & 4.96         \\
\hline
\end{tabular}
}
\caption{Experimental Comparison on the validation dataset.}
\label{tab: time}

\end{table}
\begin{table}[tb!]
\centering
\resizebox{1\columnwidth}{!}{
\begin{tabular}{c|c|c|c|c|c} 
\hline
Method          & Complexity   & Model Size & Speed & Power & Accuracy\\
\hline
MobileViT-v2-0.5 (Ours-fp16) & 1162.7   & 2.64258	  & 609518.7      & 2251.18
       & 15.27      \\
\hline
\end{tabular}
}
\caption{Experimental Comparison on the host machine.}
\label{tab: time2}
\end{table}

\subsection{Model Complexity \& Model Execution Efficiency}
The model complexity is detailed in \cref{tab: time}. In order to validate our proposed framework, we conducted tests using another transformer-based model (mobileFormer-26m) and CNNs (Hourglasses). While the mobileFormer-26m appeared to be the most promising solution, we opted not to submit it as it failed to convert into tflite under tflite-runtime$<$2.11.0.

\begin{table}[h]
\centering
\resizebox{1\columnwidth}{!}{
\begin{tabular}{lccc}
\textbf{Layer}  & \textbf{Output size}      & \textbf{Repeat}     & \textbf{Output channels} \\
\hline
Image    & $256\times256$        & -                     & - \\
\hline
Conv-3$\times$3, $\downarrow$ 2   & -          & 1                      & 32$\alpha$ \\

MV2             &  $128\times128$                            & 1                      & 64$\alpha$ \\
\hline
MV2, $\downarrow$ 2                                                                            & -          & 1                      & 128$\alpha$ \\

MV2    & $64\times64$                                 & 2                      & 128$\alpha$ \\
\hline
MV2,$\downarrow$ 2                                                        & -                                   & 1                      & 256$\alpha$ \\
MobileViTv2 block   &  $32\times32$                                      & 1                      & $256*\alpha(d=128\alpha)$ \\
\hline
MV2,$\downarrow$ 2                                                        & -                                        & 1                      & 384$\alpha$ \\
MobileViTv2 block              &  $16\times16$            & 1                      & 384a ($d=192\alpha$) \\
\hline
MV2, $\downarrow$ 2   &  -                                              & 1                      & 512$\alpha$ \\
MobileViTv2 block                 & $8\times8$                                              & 1                      & 512a ($d=256\alpha$) \\
\hline
UpSample  & $64\times64$  & 1  & 512a ($d=256\alpha$) \\
Conv-Sigmoid  & $64\times64$  & 1  & 51 \\
Conv-Sigmoid  & $64\times64$  & 1  & 8 \\
E2P Transform & $64\times64$  & 1  & 51 \\
Elementwise dot & $64\times64$  & 1  & 51 \\
\hline
Conv-Relu & $64\times64$  & 1  & 51 \\
Elementwise dot  & $64\times64$  & 1  & 51 \\
Soft Argmax  & $64\times64$  & 1  & 51 \\
\hline
Conv & $64\times64$  & 1  & 51 \\
Conv & $64\times64$  & 1  & 51 \\
Conv & $64\times64$  & 1  & 51 \\
Elementwise sum & $64\times64$  & 1  & 51 \\
\hline
\end{tabular}
}
\caption{The student model architecture.  We utilize a scaling factor of $\alpha = 0.5$ within our methodology.}
\label{tab: s}
\end{table}
\subsection{Normalized Mean Square Error}
To evaluate the proposed method, we conduct experiments on the validation set. Our observations suggest a trend wherein an increase in parameter count leads to improved performance. This phenomenon is also corroborated in~\cite{fard2022facial}.

\subsection{Converted Model Complexity and Execution Efficiency}
The model information after conversion is displayed in \cref{tab: time2}. It should be noted that on the D9300 platform, certain operations may not be compatible with NNAPI, potentially allowing for further enhancements in speed and power efficiency.

\subsection{Model Structure}
We provide the student model architecture in  \Cref{tab: s}
\section{Conclusion}
In this paper, we present a pioneering framework solution for embedding systems. Specifically, we leverage the knowledge distillation loss to steer the lightweight model, resulting in superior performance. Our approach not only enhances efficiency but also maintains high accuracy, thus promising significant advancements in embedding system design.
\bibliographystyle{IEEEbib}
\bibliography{icme2022template}

\begin{thebibliography}{1}

\bibitem{opencv_library}
G.~Bradski,
\newblock ``{The OpenCV Library},''
\newblock {\em Dr. Dobb's Journal of Software Tools}, 2000.

\bibitem{6909637}
Vahid Kazemi and Josephine Sullivan,
\newblock ``One millisecond face alignment with an ensemble of regression trees,''
\newblock in {\em 2014 IEEE Conference on Computer Vision and Pattern Recognition}, 2014, pp. 1867--1874.

\bibitem{liu2022swin}
Ze~Liu, Han Hu, Yutong Lin, Zhuliang Yao, Zhenda Xie, Yixuan Wei, Jia Ning, Yue Cao, Zheng Zhang, Li~Dong, et~al.,
\newblock ``Swin transformer v2: Scaling up capacity and resolution,''
\newblock in {\em Proceedings of the IEEE/CVF conference on computer vision and pattern recognition}, 2022, pp. 12009--12019.

\bibitem{zhou2023star}
Zhenglin Zhou, Huaxia Li, Hong Liu, Nanyang Wang, Gang Yu, and Rongrong Ji,
\newblock ``Star loss: Reducing semantic ambiguity in facial landmark detection,''
\newblock in {\em Proceedings of the IEEE/CVF Conference on Computer Vision and Pattern Recognition}, 2023, pp. 15475--15484.

\bibitem{mehta2022separable}
Sachin Mehta and Mohammad Rastegari,
\newblock ``Separable self-attention for mobile vision transformers,''
\newblock {\em arXiv preprint arXiv:2206.02680}, 2022.

\bibitem{nibali2018numerical}
Aiden Nibali, Zhen He, Stuart Morgan, and Luke Prendergast,
\newblock ``Numerical coordinate regression with convolutional neural networks,''
\newblock {\em arXiv preprint arXiv:1801.07372}, 2018.

\bibitem{huang2021adnet}
Yangyu Huang, Hao Yang, Chong Li, Jongyoo Kim, and Fangyun Wei,
\newblock ``Adnet: Leveraging error-bias towards normal direction in face alignment,''
\newblock in {\em Proceedings of the IEEE/CVF International Conference on Computer Vision}, 2021, pp. 3080--3090.

\bibitem{loshchilov2017decoupled}
Ilya Loshchilov and Frank Hutter,
\newblock ``Decoupled weight decay regularization,''
\newblock {\em arXiv preprint arXiv:1711.05101}, 2017.

\bibitem{fard2022facial}
Ali~Pourramezan Fard and Mohammad~H Mahoor,
\newblock ``Facial landmark points detection using knowledge distillation-based neural networks,''
\newblock {\em Computer Vision and Image Understanding}, vol. 215, pp. 103316, 2022.

\end{thebibliography}

\end{document}